\documentclass{ecai} 

\usepackage{eurosym}
\usepackage{latexsym}
\usepackage{amssymb}
\usepackage{amsmath}
\usepackage{amsthm}
\usepackage{booktabs}
\usepackage{enumitem}
\usepackage{graphicx}
\usepackage{color}
\usepackage{algorithm2e}
\usepackage{stmaryrd}
\usepackage{MnSymbol}
\usepackage{makecell}
\usepackage{nicefrac}

\usepackage{subcaption}

\newtheorem{theorem}{Theorem}

\def\Pref{\ensuremath{\succsim}}

\newtheorem{example}{Example}

\newcommand{\BibTeX}{B\kern-.05em{\sc i\kern-.025em b}\kern-.08em\TeX}

\begin{document}


\begin{frontmatter}


\paperid{9183} 



\title{Identifying two piecewise linear additive value functions from anonymous preference information}


\author[A, B]{\fnms{Vincent}~\snm{Auriau}\orcid{0009-0002-9640-2639}\thanks{Corresponding Author. Email: vincent.auriau@artefact.com}}
\author[A]{\fnms{Khaled}~\snm{Belahcène}\orcid{0000-0003-4502-9539}}
\author[B]{\fnms{Emmanuel}~\snm{Malherbe}\orcid{0009-0006-0898-6873}} 
\author[A]{\fnms{Vincent}~\snm{Mousseau}\orcid{0000-0001-8574-3337}}
\author[C]{\fnms{Marc}~\snm{Pirlot}\orcid{0000-0002-3689-0944}} 

\address[A]{MICS, CentraleSupélec, Université Paris Saclay, France}
\address[B]{Artefact Research Center, France}
\address[C]{MATHRO, Université de Mons, Belgique}


\begin{abstract}

Eliciting a preference model involves asking a person, named decision-maker, a series of questions. We assume that these preferences can be represented by an additive value function. In this work, we query simultaneously  two decision-makers in the aim to elicit their respective value functions. For each query we receive two answers, without noise, but without knowing which answer corresponds to which decision-maker. We propose an elicitation procedure that identifies the two preference models when the marginal value functions are piecewise linear with known breaking points.
\end{abstract}

\end{frontmatter}


\section{Introduction}
The additive value model is standard to specify preferences over a set of alternatives evaluated on several criteria, see \cite{krantz1971foundations}. More precisely, given an alternative $x$ defined by its evaluation on $n$ criteria $x=(x_1,  \ldots, x_n)$, where $x_i$ is the evaluation of $x$ on criterion $i$, the value of $x$ is defined by $u(x)=\sum_i u_i(x_i)$, where $u_i$ is the marginal value function on criterion $i$. In this paper, we consider that these marginals $u_i$ are strictly increasing and piecewise linear, with an a priori defined linear decomposition. In the following, we will call such preference model a UTA model. To elicit such a UTA model, it is usual to define standard queries corresponding to matching questions \cite{VonWinterfeldtEdwards86Book} involving two alternatives $x$ and $y$ whose evaluations differ on two criteria $i$ and $j$ only. Three evaluations out of $x_i, x_j, y_i,$ and $y_j$ are given, and the respondent provides the unfixed value in such a way that indifference holds between $x$ and $y$ (which is noted $x \sim y$). It implies that  $u_i(x_i) + u_j(x_j) = u_i(y_i) + u_j(y_j)$. The set of points $(x_i, x_j)$ in the $(i,j)$ plane for which $u_i(x_i) + u_j(x_j)$ is a constant forms an indifference curve in the $(i,j)$ plane. Using matching questions, one can identify points belonging to each indifference curve \citep{Bouyssou2016}, which will prove useful in the sequel.

We consider a framework in which we have to elicit simultaneously two UTA models. In this context, the answer to a matching query consists of two evaluations $a$ and $a'$ for the unfixed component. These responses are provided anonymously, without the ability to know to which model each answer corresponds. In this paper, our purpose is to study this problem from an identifiability perspective. In other words, we investigate whether it is always possible to define a finite sequence of queries whose answers enable the identification of the two piecewise linear additive value functions models. The identifiability problem can also be formulated as a game in which a first player specifies the queries, and a second player provides (thruthfull) answers. The question becomes: is there a finite winning strategy for the first player? In another variant of the game (not considered in this paper), the second player can answer with an (adversarial) noise. 

Although the problem is purely formulated as a formal one, it is worth noticing that it corresponds to real-world problems. For instance, a supermarket is willing to adapt the list of products to its customer base. The products selected should align with the client's preferences. Yet, not all customers necessarily have the same preferences, and one can consider a market segmentation in which each segment represents a group of clients with homogeneous preferences. 
\begin{example}
    \label{ex:Comuter-Traveler-intro}
   In this paper, we will use, as a running example, the choice among electric cars that differ on two criteria: autonomy (from 100km to 600km) and price (from 10k\euro{} to 50k\euro{}). Answers to preference queries are obtained through an online system that guarantees anonymity. Two decision-makers express their preferences: \emph{Commuter} will use the car in a urban context for relatively short distances, while \emph{Traveler} will drive long distances. \emph{Commuter} and \emph{Traveler} can use respectively 30k\euro{} and 40k\euro{} liquid savings; they need to get a loan for car that are more expensive than their savings. 
\end{example}

Such type of problem has already been tackled from a preference learning perspective \cite{Auriau-et-al-2025}, and the identifiability result proposed here is of high importance to justify the search for efficient learning algorithms that infer multiple UTA models from preference statements.

Our aim is to provide a systematic elicitation procedure to identify two additive piecewise linear value functions from a finite sequence of matching queries. In this introduction, we provide an overall presentation of this procedure; the following sections aim at providing a detailed and justified description of this elicitation procedure. Hence, the main result of the paper can be formulated as follows.

\begin{theorem}
    It is possible to elicit two piecewise linear additive preference models based on anonymized preference information using a finite number of indifference queries.
\end{theorem}

In our setting, each matching query involves alternatives that vary on two criteria $i$ and $j$ only, with three values fixed by the Eliciter and the responses are provided by the different Decision-Makers (DMs) in order to obtain an indifference statement. The value space of criteria $(i, j)$ defines a plane where the queries can be fully represented since everything else is equal. In particular, given that marginal value functions are piecewise linear, the indifference curves are also represented by (different) piecewise linear functions. In this criteria plane,  it is convenient to define two types of queries with a specific geometry that we will use as ``building blocks'' of our elicitation procedure:
\begin{itemize}
    \item \emph{Single Rectangle constraint:} The query values as well as the answers are within a single rectangle, defined by the a priori defined linear segments on both criterion scales.
    \item \emph{Neighboring Rectangles constraint:} The query values are selected such that it implies two neighboring rectangles defined by the linear segments on the criterion scales.
\end{itemize}

Single rectangle queries make it possible to elicit the anonymized pair of slopes of marginal value functions on a specific interval, while neighboring rectangles queries provide coupling constraints allowing to assign the slopes to the two value functions. 
Using these building blocks, the procedure can be summarized as follows:
\begin{itemize}
    \item Initialization: the procedure starts with the two first criteria and postulates w.l.o.g. that the slope on criterion 1 are equal to one on the first segment. An Single-Rectangle query identifies the slope of each model for the first segment on criterion 2.
    \item Iteration: given an elicited segment $\ell$ on criterion $1$, a succession of queries allows to identify the slopes of both models on segment $\ell+1$; the same applies to criterion 2.
    \item Finally, in order to obtain all marginal value functions, one can apply the same principle for all pairs of criteria (1,$i$), $i\ne 1$.
\end{itemize}

The paper is organized as follows. After a review of related works in Section 2, Section 3 provides general notation. Section 4 presents the two query types which constitute ``building blocks'' for the identification procedure. Section 5 presents results justifying the iterative nature of the general procedure presented in Section 6.
\section{Background and Related Work}
The additive value (or utility) function (AVF)  model is the most widely used and studied preference model. It may represent preference relations on sets of objects $x=(x_1, \ldots, x_n)$ described by $n$ attributes or criteria. If attribute $i$ takes values in a set $\mathcal{X}_i$ (criterion scale), let us consider a preference relation $\Pref$ on the Cartesian product $\mathcal{X} = \Pi_{i=1}^n \mathcal{X}_i$. Under well understood conditions on the preference $\Pref$ \citep{krantz1971foundations,keeneyraiffa1976,VonWinterfeldtEdwards86Book}, there exist real valued functions $u_i: \mathcal{X}_i \rightarrow \mathbb{R}$ such that, for all $x,y \in \mathcal{X}$, we have 
\begin{equation}
x\Pref y \quad \textrm{iff} \quad \sum_{i=1}^n u_i(x_i) \geq \sum_{i=1}^n u_i(y_i).
\label{def:pref-rel}
\end{equation}
The marginal value (or utility) functions $u_i$ are unique up to positive affine transformations. The main property required of $\Pref$ for admitting such a representation is an independence condition allowing for \textit{ceteris paribus} reasoning. Specifically, one may compare objects without specifying the values of the attributes on which they agree. Such characterization results allow one to devise query strategies for eliciting the marginal value functions, hence describing the preferences on the whole Cartesian product by a numerical representation. Among the 
query strategies, the method of indifference judgments builds a ``standard sequence'' of points on a criterion scale, which are equally spaced (in terms of value). This is done by using a fixed interval on the scale of another criterion as a unit measure. Queries are designed as follows. Let $q_i, p_i$ be two distinct values on the scale $\mathcal{X}_i$ of criterion $i$, and $q_j$ a value on the scale $\mathcal{X}_j$ of criterion $j$. The query is: ``What is the value $a_j \in \mathcal{X}_j$ such that the value difference between $q_i$ and $p_i$ on $\mathcal{X}_i$ is exactly compensated by the value difference between $q_j$ and $a_j$ on $\mathcal{X}_j$ (all other things being equal, i.e., ceteris paribus)?'' The answer $a_j$ is a first point in a standard sequence. The next point is the answer to the modified query obtained by substituting $q_j$ by $a_j$ in the initial query, and so on. The pairs $(q_i, q_j)$, $(p_i, a_j)$ belong to an indifference curve in the domain $\mathcal{X}_i \times \mathcal{X}_j$ (hence the name ``indifference judgments'').   

Jacquet-Lagr\`{e}ze and Siskos \cite{jaquetlsiskos1982} developed a method, called UTA, to learn marginal value functions of an AVF that is compatible with known preferences expressed on a set of pairs of objects. In order to exploit linear programming they assume that the marginal value functions are piecewise linear. Let the points $x_{i,\ell}, \ell = 0, 1, \ldots, L$ partition the scale $\mathcal{X}_i$ of criterion $i$ into $L$ subintervals $[x_{i,\ell}, x_{i,\ell+1}]$, and the marginal value function be piecewise linear with respect to this decomposition. The latter is known as soon as the values $u_i(x_{i,\ell})$ at the intervals endpoints are determined. Alternatively, it suffices to determine the slopes of the linear segments of the marginal value function $\gamma_{i,\ell} = [u_i(x_{i,\ell}) -u_i(x_{i,\ell-1})] / (x_{i,\ell} -x_{i,\ell-1})$.  
This method has been subsequently widely developed and applied \citep{jaquetlsiskos2001}. In general, there are several models (actually, an infinite number of models) compatible with the set of preference examples. Some papers \citep[e.g.,][]{GrecoMousseauSlowinski08} deal with formulating robust conclusions by  considering the set of all models compatible with the examples. 

In this work, we consider the case in which two DMs reply to elicitation queries each using her own AVF model, with piecewise linear marginals. The answers to the queries are not tagged by the name of the DM. The question we tackle is whether it is possible to identify each DM's model without ambiguity, by asking queries of the type ``indifference judgments''. 

Similar work about model identifiability problems have been addressed in multicriteria decision analysis in the case of the Choquet integral \cite{Labreuche_Hüllermeier_etal2016}, the hierarchical Choquet integral \cite{KR2021-15} and more \cite{braziunas2006preference,regan2011eliciting}.  
This question of model identifiability can be approached from different angles. At its core is a parametric family of functions $(f_\omega)_{\omega\in \Omega}$ where $\Omega$ is the latent parameter set, and each $f_\omega$ is a function mapping queries in $\mathcal Q$ to answers in $\mathcal A$. Abstract identifiability can be cast as the existence of a query $q\in \mathcal Q$ such that two distinct parameters $\omega\ne\omega'\in\Omega$ yield distinct outcomes, i.e. $f_\omega(q)\neq f_{\omega'}(q)$. More operational definitions exist. For instance, various notions of combinatorial dimension have been put forward so as to upper bound the number of queries needed to distinguish concepts \cite{ANGLUIN2004175}. In turn, these bounds can be used to obtain statistical guarantees w.r.t. the estimation of those parameters from data, playing a key role in probably approximately correct machine learning (see e.g. \cite{SSS,Ran2017}).
The problem we address here is not standard w.r.t. those approaches, because neither the queries nor the latent parameters are. Indifference queries do not have a binary outcome, as expected from concept learning theory. More importantly, our latent parameter space encompasses both the preference models of the DMs but also the identity of the DM associated to each answer (and thus grows with the number of queries). A more fruitful analogy comes from adversarial game-playing: consider two players, the Eliciter and the Adversary, and a fixed parameter space, such that $\omega$ completely specifies two UTA models. The Eliciter can either query the Adversary, or reveal two UTA models. The Adversary answers queries as he wishes. He immediately loses if there are no parameter $\omega$ compatible with all answers, and wins if the Eliciter reveals a parameter $\omega_E$ and he can provide an alternate $\omega'\ne\omega$ which is compatible with all answers. A procedure based on mixed-integer linear programming is described in \cite{Auriau-et-al-2025} and allows to decide whether a set of preference statements can be partitioned into two DMs represented by a UTA model, and can serve as a judge in this game. In effect, this allows the Adversary to cheat and modify the model behind the scene, as long as this behavior cannot be detected, and forces the Eliciter to reduce the set of all possible models to a singleton. Such approach can be found in \cite{Lagrue2018}. Here, we show that the Eliciter has indeed a winning strategy.

\section{Problem setting}
We consider a set $\mathcal{N}=\{ 1,..., n \}$ of criteria. A criterion $i \in \mathcal{N}$ is defined on its scale, noted $\mathcal{X}_i \subset \mathbb{R}$. For each criterion $i$, we are given a positive integer $L_i$ and $L_i+1$ values $x_{i,0} < x_{i,1} < \dots < x_{i,L_i}\in\mathcal X_i$.

The value function of a DM $\kappa$ in the form of a UTA model can be defined by its values at the edges of the intervals, $(u^\kappa_{i}(x_{i, \ell}))_{\ell \leq L_i, i \in \mathcal{N}}$. An equivalent formulation uses the slopes of this value function $\gamma_{i, \ell}^\kappa = \frac{u_{i}^\kappa(x_{i, \ell}) - u_{i}^\kappa(x_{i, \ell-1})}{x_{i, \ell}-x_{i, \ell-1}}$ for $1\leq \ell \leq L_i$. We consider the functions to be strictly increasing, meaning that $\gamma^\kappa_{i, \ell} > 0$. 

\begin{figure*}
\centering
\begin{subfigure}[B]{.31\textwidth}
  \centering
  \includegraphics[width=1.\linewidth]{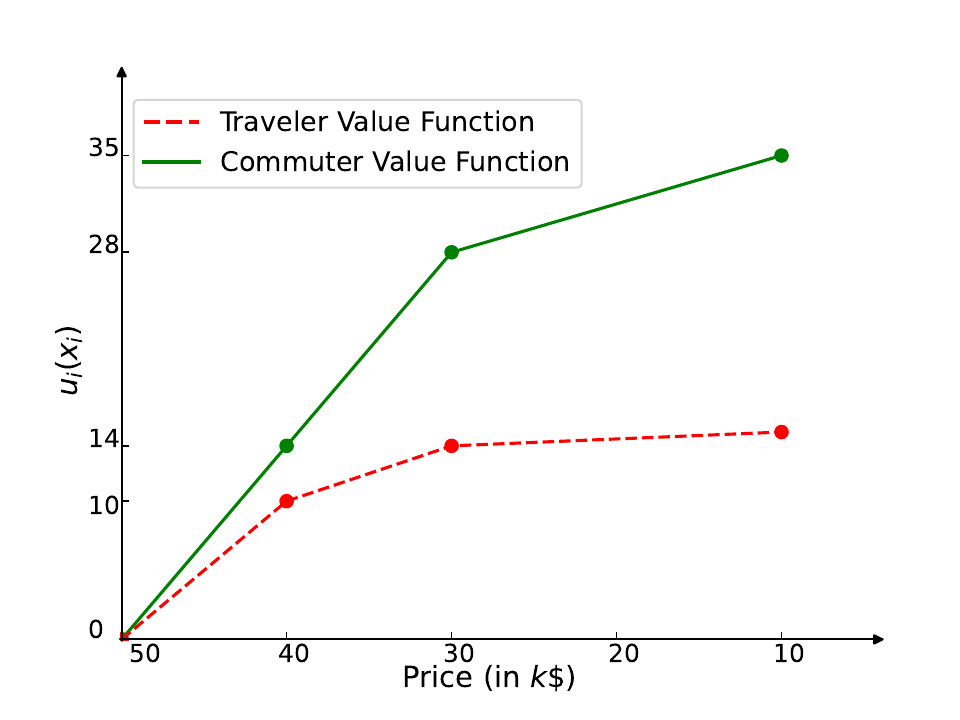}\vskip-1ex
  \caption{\footnotesize Marginal value of Price}
  \vskip3ex
\end{subfigure}%
\begin{subfigure}[B]{.31\textwidth}
  \centering
  \includegraphics[width=1.\linewidth]{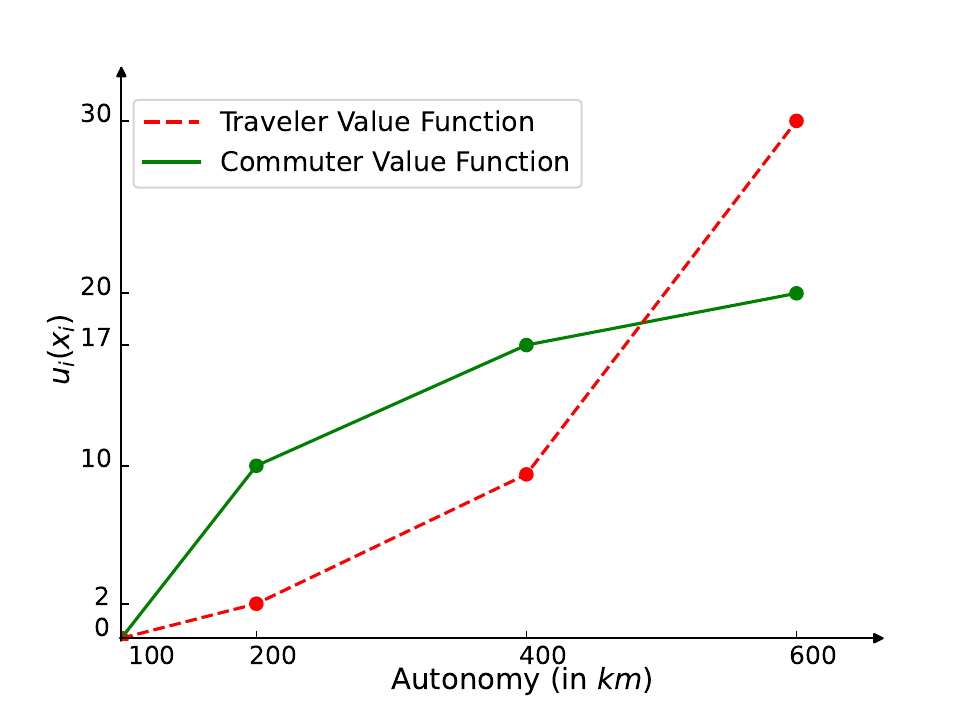}\vskip-1ex
  \caption{\footnotesize Marginal value of Autonomy}
  \vskip3ex
\end{subfigure}
\begin{subfigure}[B]{.31\textwidth}
  \centering
  \includegraphics[width=1.\linewidth]{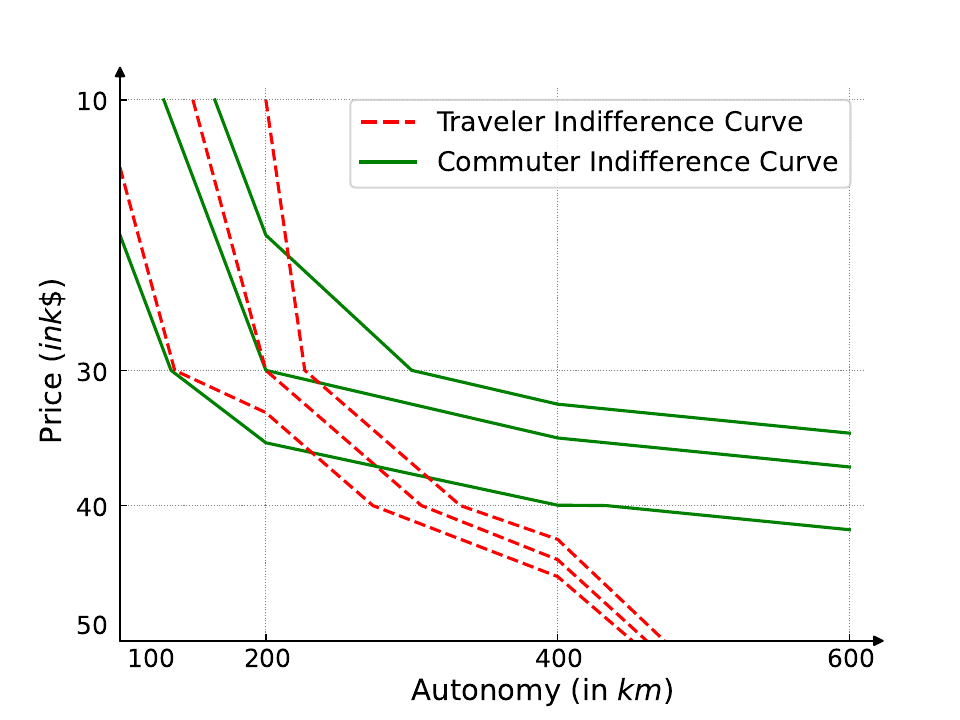}\vskip-1ex
  \caption{\footnotesize Indifference curves for Price vs Autonomy}
  \vskip3ex
\end{subfigure}
\caption{Preferences of the DMs from the example, serving as ground truth for the elicitation. Marginal values for Price (left), Autonomy (middle), and resulting indifference curves (right).
\\
}
  \label{fig:ex_ex}
\end{figure*}

\begin{example}
    \label{ex:Comuter-Traveler-sec3-1}
    (Ex. \ref{ex:Comuter-Traveler-intro} cont.) The marginal value functions for both DMs (Traveler \& Commuter) are provided in Figure \ref{fig:ex_ex}(a) for the price criterion, and in Figure \ref{fig:ex_ex}(b) for the autonomy criterion. These value functions induce interpretable preferences: for instance, an increase in autonomy from 100km to 200km is valued 2 (10, resp.) for \emph{Traveler} (for  \emph{Commuter}, resp.). Indeed, 200km autonomy is a poor value for  \emph{Traveler} who drives long distances, while it is already reasonably good for \emph{Commuter} who drives shorter distances.
\end{example}

Given the value function $u^\kappa$ of DM $\kappa$, one can define \emph{Indifference curves} which are the sets of points $x\ \in \prod_{i \in {\cal N}}{\cal X}$ in the Cartesian product of criteria having the same value, i.e., such that $u^\kappa(x) = z \; ( z \in \mathbb{R})$. All pairs of objects on an Indifference Curve are said to be indifferent, meaning they belong to the symmetric part $\sim$ of the $\succsim$ relation defined in (\ref{def:pref-rel}). 

\begin{example}
    \label{ex:Comuter-Traveler-sec3-2}
    (Ex. \ref{ex:Comuter-Traveler-sec3-1} cont.) Indifference curves induced by the value functions of \emph{Traveler} and \emph{Commuter} are depicted in Figure \ref{fig:ex_ex}(c). Obviously, for a given DM, indifference curves do not cross. On the contrary the indifference curves of the different DMs do cross, meaning their preferences are different.
\end{example}

One can note that the criteria space $\mathcal{X}_i \times \mathcal{X}_j$ can be seen as a grid where each rectangle element is defined by the breaking points on each criterion: $\left(x_{i, \ell_i-1}, x_{j, \ell_j-1} \right); \left(x_{i, \ell_i}, x_{j, \ell_j} \right)$. In this specific rectangle denoted $\mathcal R_{\ell_i,\ell_j}$ (corresponding to the $\ell_i^{th}$ interval of criterion $i$ and the the $\ell_j^{th}$ interval of criterion $j$), the indifference represented by the value functions is a straight line, neither horizontal nor vertical, whose slope is  $\gamma^\kappa_{i, \ell_i}/ \gamma^\kappa_{j, \ell_j}>0$.


We consider two DMs, noted $\alpha$ and $\beta$, that can be queried at will. We define a query $\mathcal{Q}$ as questioning the DMs concerning alternatives whose evaluations vary on two criteria only. More specifically, we formulate $\mathcal{Q}$ as the specification of an ordered pair of criteria, $(i, j) \in \mathcal{N}^2$ and a triplet of values on these criteria: $(q_{i}, q_{j}, p_{i})$, with $(q_{i}, p_{i}) \in \mathcal{X}_i^2$ and $q_{j} \in \mathcal{X}_j$. This query triggers the collection of answers $\mathcal{A} = \{ a^{\kappa}_{j}, a^{\kappa'}_{j}\} \subset \mathcal X_j$ from the two DMs. We arbitrarily decide to note $\kappa$ and $\kappa'$ such that $a^{\kappa}_{j} \leq a^{\kappa'}_{j}$. We also sometimes omit to specify the selected criteria, when it is already clear enough. The elements  of $\mathcal{A}$  are such that either $\left[(q_{i}, q_{j}) \sim_\alpha (p_{i}, a_{j}^{\kappa}), (q_{i}, q_{j}) \sim_\beta (p_{i}, a_{j}^{\kappa'}) \right]$ or $\left[(q_{i},  q_{j}) \sim_\beta (p_{i}, a_{j}^{\kappa}), (q_{i}, q_{j}) \sim_\alpha (p_{i}, a_{j}^{\kappa'}) \right]$, with $\sim_{\alpha}$ (resp. $\sim_{\beta}$) denoting indifference for DM $\alpha$ (resp. $\beta$). In other words, the answers are non-identifiable, meaning that we don't known which DM they come from.Some queries cannot be answered by one, or both DMs, when the difference of value on the first criterion is too large to be compensated on the second one. In this case, we are provided with the answer ``None''.

\begin{example}
    \label{ex:Comuter-Traveler-sec3-3}
    (Ex. \ref{ex:Comuter-Traveler-sec3-2} cont.) a query $Q$=(200km, 30k\euro{}) $\sim$ (400km, ?) can be interpreted as: given a car with a 30k\euro{} price having 200km autonomy, how much more would you be willing to pay to obtain 400km autonomy? the two answers are (34k\euro{}, 43k\euro{}) without knowing which DM gave each answer.
\end{example}

Eliciting the value function of both DMs involves a sequence of queries, and for each of these queries, there is a hidden bit of information. Hence, the problem of attributing each answer to the correct DM compounds at each step--giving a combinatorial aspect, potentially explosive, to this problem. A summary of the notation can be found in Table \ref{tab:notations}.

\label{sec:uta}

\begin{table}[]
    \caption{Notations summary}
    \label{tab:notations}
    \centering
    \begin{tabular}{lll}
    \toprule
       Description & Notation & \\
       \midrule
       Criteria  set & $\mathcal{N} = \{1,..., n \}$ & $i, j \in \mathcal{N}$\\
       Criterion scale & $\mathcal{X}_i = [x_{i, 0}, x_{i, L}]$ & \\
       Criterion interval & $[x_{i ,\ell}, x_{i, \ell+1}]$ & $\ell \in \{0, ..., L\}$\\
        Decision makers & $\alpha, \beta$ &\\
        Value functions & $u^\kappa(\cdot)$ & $\kappa \in \{\alpha, \beta \}$\\
        \makecell[l]{Marginal value\\ functions} & $u^\kappa_i(\cdot)$ & $\kappa \in \{\alpha, \beta \}, i \in \mathcal{N}$\\
       \makecell[l]{ Value function\\ slopes} & $\gamma^\kappa_{i, \ell}$ & \makecell[l]{$\kappa \in \{\alpha, \beta \}, i \in \mathcal{N},$\\$ 1 \leq \ell \leq L$}\\
        Queries & $\mathcal{Q}= (q_i, q_j) \sim (p_i, ?)$ & $q_i \in \mathcal{X}_i, p_j \in \mathcal{X}_j$\\
        Answers & $a_i^{\kappa}, a_i^{\kappa'}$ & $i\in \mathcal{N}, a_i^{\kappa} \in \mathcal{X}_i$\\ \bottomrule
        
    \end{tabular}
\end{table}



\section{Play Patterns}
\label{sec:PP}

Our perfect Eliciter relies on two basic play patterns or sequences of moves (i.e. indifference queries) yielding preference information that allows it to reach specific short-term goals. Given two criteria $i\ne j \in \mathcal N$, these goals can be specified geometrically in the plane $\mathcal X_i \times \mathcal X_j$: obtaining indifference statements between points located either in the same rectangle (Section \ref{sub:SRPI}) or adjacent rectangles (Section \ref{sub:ARPI}). 


\subsection{Obtaining Single-Rectangle Preference Information}
\label{sub:SRPI}

We are first interested in collecting preference information in the form of anonymized indifference statements between points lying in the same rectangle $\mathcal R_{\ell_i+1,\ell_j+1}$. Play Pattern \ref{alg:SRC} ensures this information can be obtained in at most two queries.

\begin{figure}[h]
    \centering
 \includegraphics[width=0.45\textwidth]{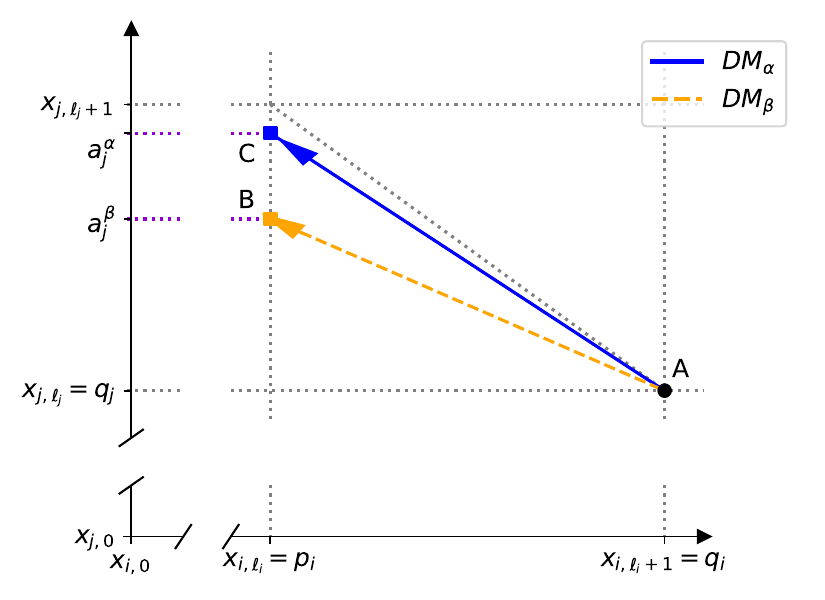}
\caption{A successful Single Rectangle query in the space $\mathcal{X}_i \times \mathcal{X}_j$, corresponding to case 1. Colored lines represent DMs' indifference curves.\\}
\label{fig:qta_i}
\end{figure}

\begin{figure}
\centering
\begin{subfigure}{.48\linewidth}
  \centering
  \includegraphics[width=1.\linewidth]{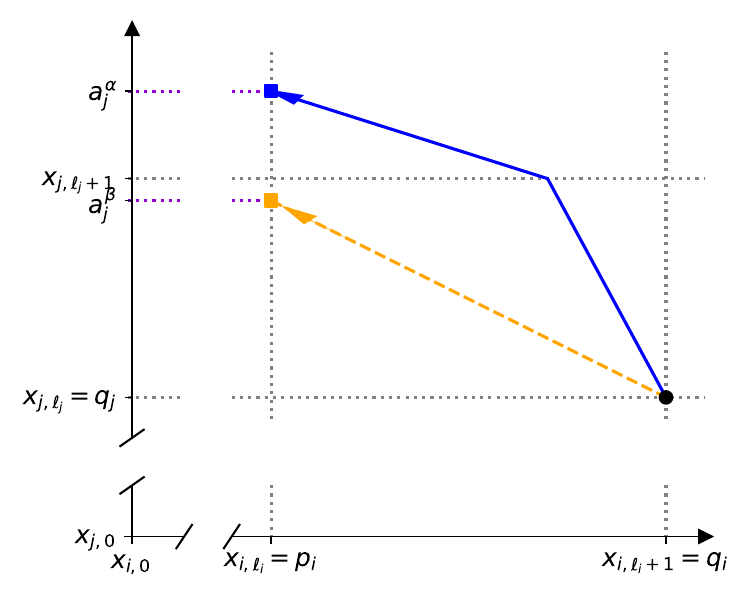}
\end{subfigure}%
\begin{subfigure}{.48\linewidth}
  \centering
  \includegraphics[width=1.\linewidth]{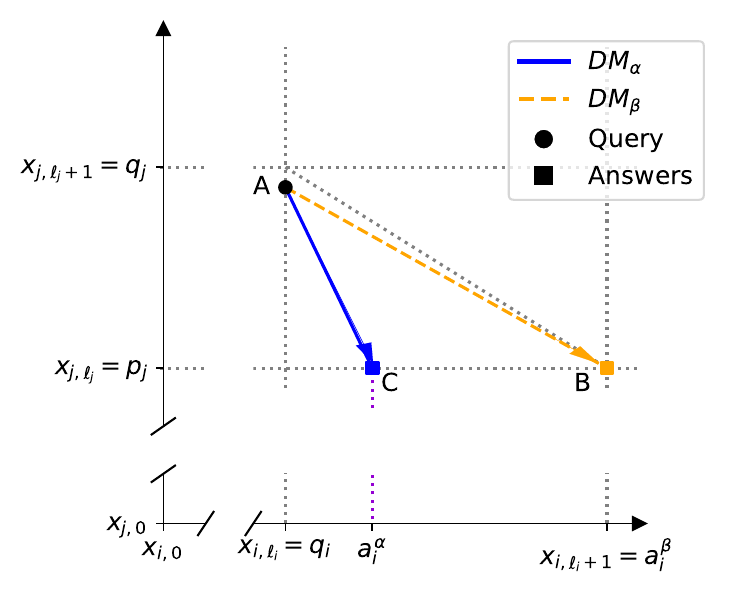}
  
\end{subfigure}
\caption{
Outcome and second query for case 2 of the Single Rectangle Constraint.\\
\\}
  \label{fig:qta_iii}
\end{figure}

\begin{figure}
\centering
\begin{subfigure}{.47\linewidth}
  \centering
  \includegraphics[width=1.\linewidth]{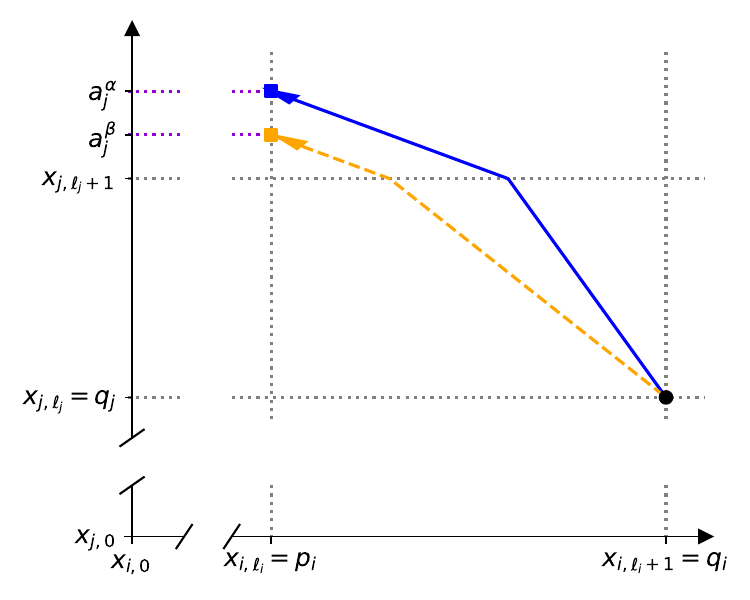}
\end{subfigure}%
\begin{subfigure}{.52\linewidth}
  \centering
  \includegraphics[width=1.\linewidth]{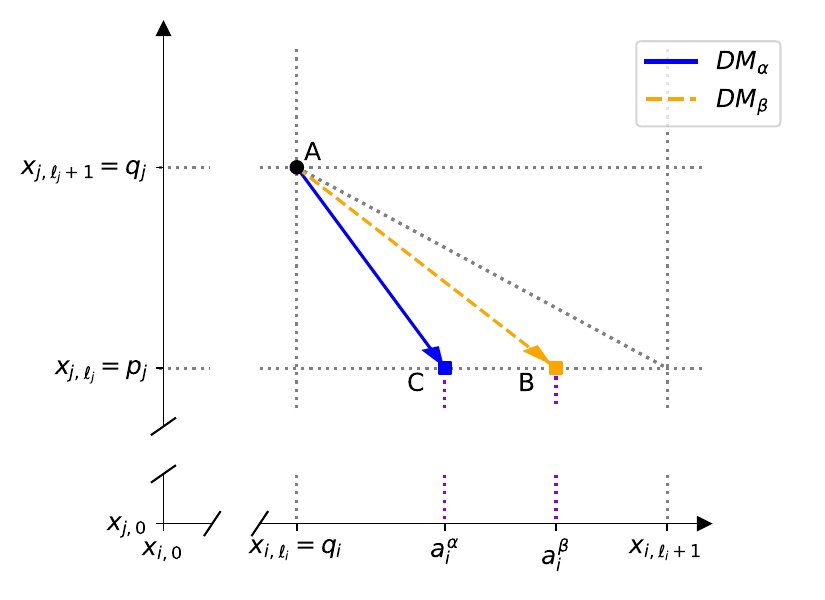}
  
\end{subfigure}
\caption{
Outcome and second query for case 3 of the Single Rectangle Constraint.}
  \label{fig:qta_ii}
\end{figure}

The play pattern opens with the following query from the bottom right corner to the left side: 
$(i:x_{i,\ell_i+1},j:x_{j,\ell_j}) \sim (i:x_{i,\ell_i},j:?),$
yielding two answers $a^1 < a^2 \in \mathcal X_i\cup\{\text{None}\}$. There are three cases:
\begin{itemize}
\item If $x_{j,\ell_j} \le a^1 \le a^2 \le x_{j,\ell_j+1}$ (Case 1, illustrated by Fig. \ref{fig:qta_i}). Both indifference curves lie below the diagonal and hit the left side of the rectangle. The specified goal is reached in a single query.
\item Else, if $a^1 \le x_{j,\ell_j}$ (Case 2, illustrated by Fig. \ref{fig:qta_iii}). Exactly one DM has an indifference curve (in blue) steeper than the diagonal and that hits the top side of the rectangle. Querying on the bottom side from the point given by the other DM (in yellow), makes sure to obtain two values $b^1 < b^2\in \mathcal X_j$ such that $b^2=x_{j,\ell_j+1}$, yielding two points on the bottom side of the rectangle.
\item Finally, Case 3 covers all situations where both indifference curves are steeper than the diagonal and hit the top side of the rectangle, as illustrated by Fig. \ref{fig:qta_ii}. Querying from the top left corner ensures getting both answers on the bottom side of the rectangle.
\end{itemize}

\RestyleAlgo{ruled}

\SetKwComment{Comment}{/* }{}
\SetKwBlock{initz}{\textbf{Initialize criterion 1:}}{}
\SetKwBlock{init}{\textbf{Initialize criterion 2: }}{}
\SetKwBlock{initi}{\textbf{Initialize criterion $i$: }}{}
\SetKwBlock{query}{\textbf{Query: }}{}
\SetKwBlock{eliciti}{\textbf{Fully Elicit Criterion $i$: }}{}
\SetKwBlock{elicitundeux}{\textbf{Fully Elicit Criterion 1, 2: }}{}
\SetKwFor{For}{for}{}{}
\renewcommand*{\algorithmcfname}{Play Pattern}

\begin{algorithm}
    \caption{to get a Single Rectangle Constraint}\label{alg:SRC}
    \KwIn{two criteria $i\ne j$, two interval labels $\ell_i\in\llbracket 0, L_i-1\rrbracket$, $\ell_j\in\llbracket 0, L_j-1\rrbracket$ }
    \KwOut{three points $A,B\neq A,C\neq A$ in $[x_{i,\ell_i},x_{i,\ell_i+1}] \times$ $[x_{j,\ell_j},x_{j,\ell_j+1}]$ s.t. one DM is indifferent between $A$ and $B$ and the other is indifferent between $A$ and $C$ }
    \BlankLine
    $a^1,a^2 \leftarrow \text{Query}\ (i:x_{i,\ell_i+1},j:x_{j,\ell_j}) \sim (i:x_{i,\ell_i},j:?)$\;
    \uIf(\Comment*[f]{Case 1}){$a^2 \le x_{j,\ell_j+1}$}{\Return{$(x_{i,\ell_i+1},x_{j,\ell_j}), (x_{i,\ell_i},a^1), (x_{i,\ell_i},a^2)$\;}}
    \uIf(\Comment*[f]{Case 2}){$a^1 \le x_{j,\ell_j+1}$}{$b^1,b^2 \leftarrow \text{Query}\ (i:x_{i,\ell_i},j:a^1) \sim (j:x_{j,\ell_j},i:?)$\;
    \Return{$(x_{i,\ell_i},a^1), (b^1,x_{j,\ell_j}), (b^2,x_{j,\ell_j})$}\;}
    \uElse(\Comment*[f]{Case 3}){$b^1,b^2 \leftarrow \text{Query}\ (i:x_{i,\ell_i},j:x_{j,\ell_j+1}) \sim (j:x_{j,\ell_j},i:?)$\;
    \Return{$(x_{i,\ell_i},x_{j,\ell_j+1}), (b^1,x_{j,\ell_j}), (b^2,x_{j,\ell_j})$}\;}

\end{algorithm}

\paragraph{Exploiting Single-Rectangle Preference Information}
Suppose that we have three points in $[x_{i,\ell_i},x_{i,\ell_i+1}] \times$ $[x_{j,\ell_j},x_{j,\ell_j+1}]$ $A (A_i,A_j)$, $B (B_i,B_j)\neq A$, $C (C_i,C_j)\neq A$  such that one DM --say $\kappa\in\{\alpha,\beta\}$-- is indifferent between $A$ and $B$ and the other --say $\kappa'\ne\kappa\in\{\alpha,\beta\}$-- is indifferent between $A$ and $C$. 

\begin{align*}
    \left\{
\begin{aligned}
    (A_i, A_j)\sim_{\kappa} (B_i, B_j)\\
    (A_i, A_j)\sim_{\kappa'} (C_i, C_j)\\
\end{aligned}
    \right.
\Leftrightarrow &
    \left\{
\begin{aligned}
    u_{\kappa}((A_i, A_j)) = u_{\kappa}((B_i, B_j))\\
    u_{\kappa'}((A_i, A_j)) = u_{\kappa'}((C_i, C_j))
\end{aligned}
    \right.
\end{align*}
The additive value function is such that
$u^\kappa((A_i, A_j)) = u_{i}^\kappa(x_{i, \ell_i}) + \gamma^\kappa_{i, \ell_i+1} \cdot (A_i - x_{i, \ell_i}) +  u^\kappa_{j}(x_{j, \ell_j}) + \gamma^\kappa_{j, \ell_j+1} \cdot (A_j - x_{j, \ell_j})$, with a similar form for $u^{\kappa'}$. More details are shared in Appendix \ref{app:eq_2}. It leads to:
\begin{align*}
    \left\{
\begin{aligned}
    \gamma^\kappa_{j, \ell_j} = \Lambda_{\kappa} \cdot \gamma^\kappa_{i, \ell_i}\\
    \gamma^{\kappa'}_{j, \ell_j} = \Lambda_{\kappa'} \cdot \gamma^{\kappa'}_{i, \ell_i}\\
\end{aligned}
    \right.  \text{\hspace{.4cm} with: } 
    \left\{
\begin{aligned}
    \Lambda_{\kappa} = \frac{B_i - A_i}{A_j - B_j}\\
    \Lambda_{\kappa'} = \frac{C_i - A_i}{A_j - C_j}\\
\end{aligned}
    \right.
\end{align*}

If $B=C$ then $\gamma^\alpha_{i}/\gamma_{j}^\alpha = \gamma_{i}^\beta /\gamma^\beta_{j}$. Otherwise, let $k\in\{0,1\}$ such that $k=0$ iff $\alpha$ answers $B$ and $\beta$ answers $C$. We have:

\begin{equation}
        \left\{
    \begin{aligned}
    \gamma^\alpha_{j, \ell_j} = \left( k \cdot \Lambda_{\kappa}  + (1-k) \cdot \Lambda_{\kappa'} \right) \cdot \gamma^\alpha_{i, \ell_i}\\
    \gamma^\beta_{j, \ell_j} = \left( (1-k) \cdot \Lambda_{\kappa}  + k \cdot \Lambda_{\kappa'} \right) \cdot \gamma_{i, \ell_i}^\beta\\
    \end{aligned}
    \label{eq:k_t}
        \right.
\end{equation}

\subsection{Obtaining Neighboring-Rectangles Preference Information}
\label{sub:ARPI}

We are also interested in collecting preference information in the form of anonymized indifference statements $A\sim_\kappa B$, $A\sim_{\kappa'} C$ between points lying in adjacent rectangles, i.e. $A \in \mathcal R_{\ell_i,\ell_j-1}$ and $B,C  \in \mathcal R_{\ell_i,\ell_j+1}$. Play Pattern \ref{alg:NRC} ensures this information can be obtained with a finite number of queries.

\begin{figure}[h]
    \centering
 \includegraphics[width=0.45\textwidth]{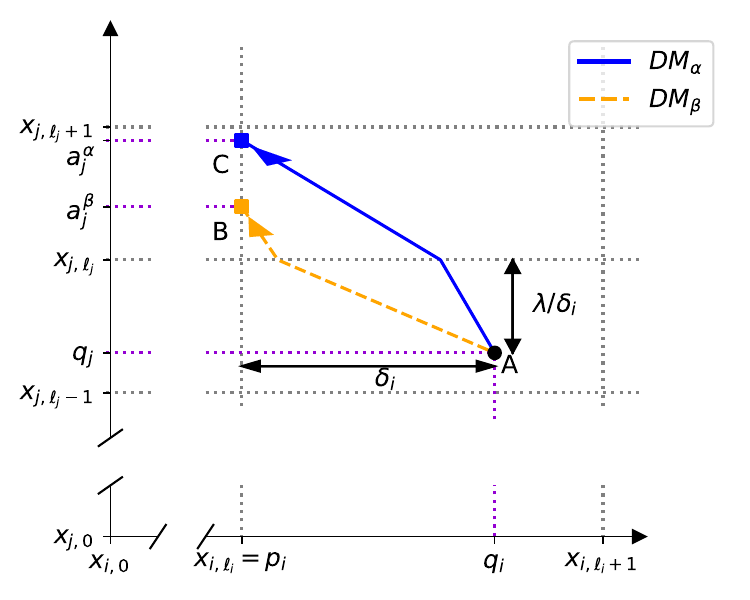}
\caption{Example of a Query with a neighboring rectangles constraint. Colored lines represent the indifference curve of each DM.\\
\\}
\label{fig:qsad_1}
\end{figure}

\begin{figure}
\centering
\begin{subfigure}{.48\linewidth}
  \centering
  \includegraphics[width=1.\linewidth]{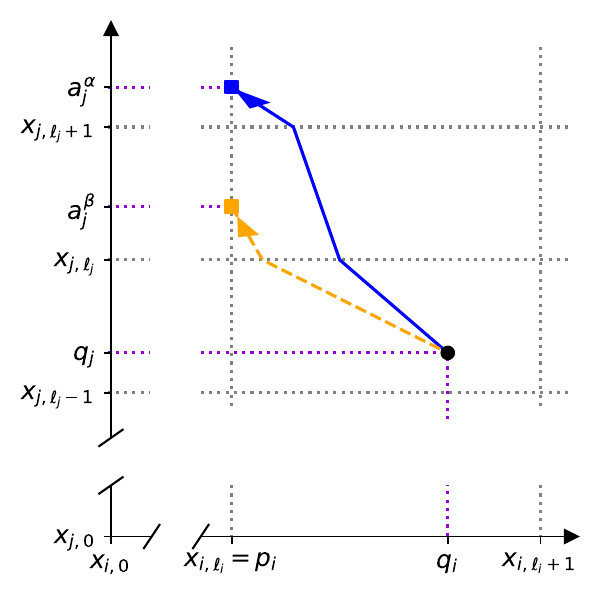}
\end{subfigure}%
\begin{subfigure}{.48\linewidth}
  \centering
  \includegraphics[width=1.\linewidth]{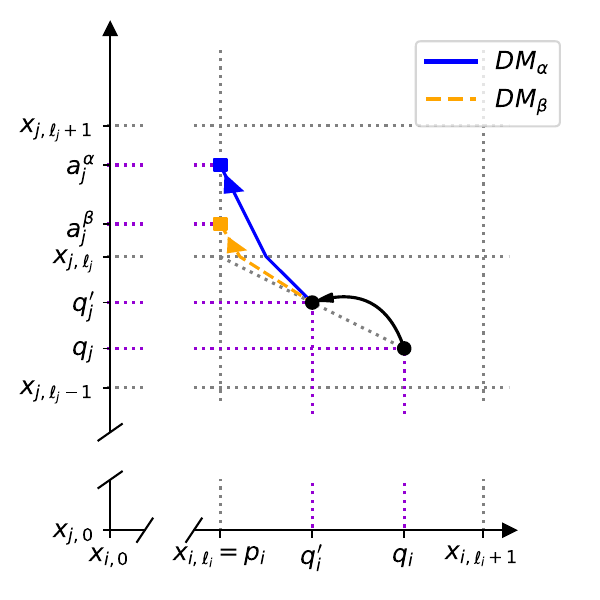}
  
\end{subfigure}
\caption{
Outcome and second query for case 1 of Play Pattern \ref{alg:NRC}.} \vskip4ex

  \label{fig:pp2_c2}
\end{figure}

\begin{figure}
\centering
\begin{subfigure}{.48\linewidth}
  \centering
  \includegraphics[width=1.\linewidth]{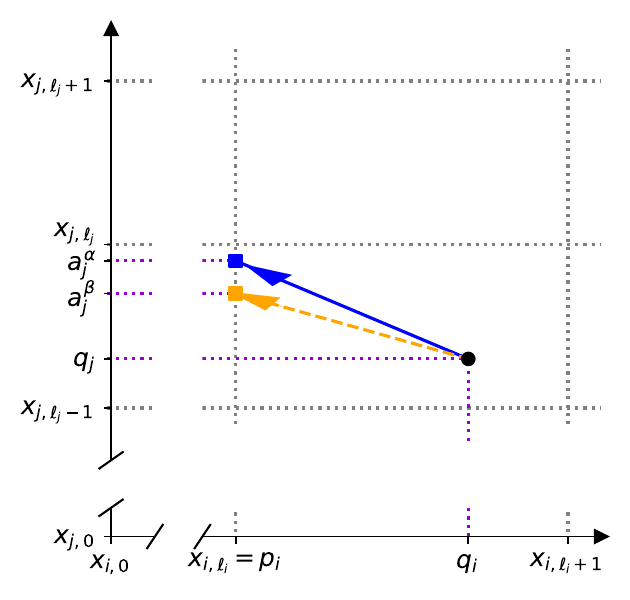}
\end{subfigure}%
\begin{subfigure}{.48\linewidth}
  \centering
  \includegraphics[width=1.\linewidth]{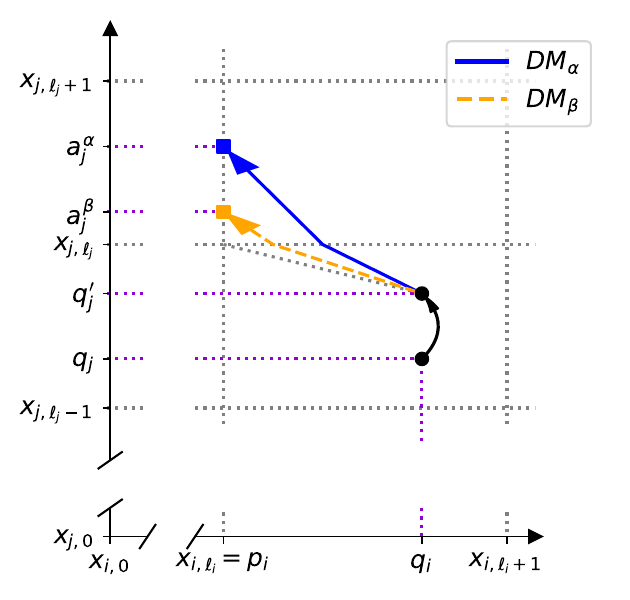}
  
\end{subfigure}
\caption{
Outcome and second query for case 2 of Play Pattern \ref{alg:NRC}.\\}
  \label{fig:qsad_3}
\end{figure}

The play pattern is based on the iterated uses of queries of the form $(i:x_{i,\ell_i}+\delta_i,j:\frac 1 2 (x_{j,\ell_j-1}+x_{j,\ell_j})+\frac \lambda 2 \delta_i) \sim (i:x_{i,\ell_i},j:?)$. The values $\lambda$ and $\delta_i$ are initialized so that the first query point lies at the center of rectangle $\mathcal R_{\ell_i,\ell_j}$. These values are then adjusted according to the answers until both answers are in the sought rectangle and the pattern terminates, as illustrated by Figure \ref{fig:qsad_1}. Case 1 corresponds to the situation where both answers are above $x_{j,\ell_j}$ but at least one is not below $x_{j,\ell_j+1}$. In such a case, $\delta_i$ is divided by two and the query point moves left (see Figure~\ref{fig:pp2_c2})- along a line passing through the point $(x_{i,\ell_i},x_{j,\ell_j})$ whose slope is controlled by $\lambda$. The variable $\lambda$ is initialized so that this line coincides with the diagonal of the lower rectangle. If, at the first query, the lowest answer is below $x_{j,\ell_j}$ we obtain single-rectangle preference information for one of the DMs and can compute the actual value of the slope of its indifference curve, which is stored in $\lambda$ (Case 2, illustrated by Figure \ref{fig:qsad_3}). In effect, the querying point moves towards the upper-left vertex of the lower rectangle along a more gentle slope, ensuring Case 2 never occurs again.




\begin{algorithm}
    
    \caption{to get a Neighboring Rectangles Constraint}\label{alg:NRC}
    \KwIn{two criteria $i\ne j$, two interval labels $\ell_i\in\llbracket 1, L_i \rrbracket$, $\ell_j\in\llbracket 1, L_j - 1\rrbracket$ 
    }
    \KwOut{three points $A,B,C$ with $A\in[x_{i,\ell_i},x_{i,\ell_i+1}] \times$ $[x_{j,\ell_j-1},x_{j,\ell_j}]$ and $B,C\in [x_{i,\ell_i},x_{i,\ell_i+1}] \times$ $[x_{j,\ell_j},x_{j,\ell_j+1}]$ such that one DM is indifferent between $A$ and $B$ and the other is indifferent between $A$ and $C$ }
    \BlankLine
    $\delta_i \leftarrow \frac{1}{2} \cdot (x_{i, \ell_{i+1}} - x_{i, \ell_i})$\;\vskip.5ex
    $\lambda \leftarrow (x_{j,\ell_j}-x_{j,\ell_j-1}) / (2\delta_i)$ \;
    $a^1,a^2 \leftarrow \text{Query}\ (x_{i,\ell_i}+\delta_i, x_{j,\ell_j-1}-\lambda \cdot \delta_i) \sim (x_{i,\ell_i},?)$\;
    \While{not $(a^1 > x_{j,\ell_j}$ and $a^2 \le x_{j,\ell_j+1})$}{
    \lIf(\Comment*[f]{Case 1}){$a^1 > x_{j,\ell_j}$}{$\delta_i \leftarrow \delta_i/2$}
    \lElse(\Comment*[f]{Case 2}){$\lambda \leftarrow {(x_{j,\ell_j}-a^1)} / {(2\delta_i)}$}
    
    $a^1,a^2 \leftarrow \text{Query}\ (x_{i,\ell_i}+\delta_i,x_{j,\ell_j}- \lambda \cdot \delta_i) \sim (x_{i,\ell_i},?)$\;}
    
    \Return{$(x_{i,\ell_i}+\delta_i,x_{j,\ell_j}-  \lambda \cdot \delta_i)$, $(x_{i,\ell_i},a^1)$, $(x_{i,\ell_i},a^2)$}\;
\end{algorithm}

\paragraph{Exploiting Neighboring-Rectangles Preference Information.}
Suppose that we have $A (A_i,A_j)$ in $[x_{i,\ell_i},x_{i,\ell_i+1}] \times$ $[x_{j,\ell_j-1},x_{j,\ell_j}]$ and $B (B_i,B_j)\neq A$, $C (C_i,C_j)\neq A$ in $[x_{i,\ell_i},x_{i,\ell_i+1}] \times$ $[x_{j,\ell_j},x_{j,\ell_j+1}]$ such that one DM is indifferent between $A$ and $B$ and the other, is indifferent between $A$ and $C$. At this point, we attribute $B$ to the DM $\kappa$ and $C$ to $\kappa'$, but we do not know whether $(\kappa,\kappa')$ is $(\alpha,\beta)$ or $(\beta,\alpha)$.

\begin{align*}
    \left\{
\begin{aligned}
    (A_i, A_j)\sim_{\kappa} (B_i, B_j)\\
    (A_i, A_j)\sim_{\kappa'} (C_i, C_j)\\
\end{aligned}
    \right.
\end{align*}
Contrary to the Single Rectangle, we have: $u_{j}^\kappa(B_j) - u_{j}^\kappa(A_j) = \gamma_{j, \ell_j+1}^\kappa\cdot (B_j- x_{j, \ell_j}) + \gamma^\kappa_{j, \ell_j}\cdot(x_{j, \ell_j}-A_j)$, leading to:
\begin{align*}
&
    \left\{
\begin{aligned}
    &\gamma_{j, \ell_j+1}^\kappa = \gamma_{i, \ell_i}^\kappa \cdot \Theta_\kappa + \gamma_{k, \ell_j}^\kappa \cdot \Phi_\kappa  \\
    &\gamma_{j, \ell_j+1}^{\kappa'} = \gamma_{i, \ell_i}^{\kappa'} \cdot \Theta_\kappa' + \gamma_{j, \ell_j}^{\kappa'} \cdot \Phi_{\kappa'}  \\
\end{aligned}
\right.
\end{align*}
\begin{equation*}
   \text{with:}\; \; \left\{
    \begin{aligned}
        &\Theta_\kappa = \frac{A_i - B_i}{B_j - x_{j, \ell_j}}; \; \Theta_{\kappa'} = \frac{A_i - C_i}{C_j - x_{j, \ell_j}}\\
        &\Phi_{\kappa} = \frac{A_j - x_{j, \ell_j}}{B_j - x_{j, \ell_j}}; \;  \Phi_{\kappa'} = \frac{A_j - x_{j, \ell_j}}{C_j - x_{j, \ell_j}}\\
    \end{aligned}
        \right.\\
\end{equation*}

Let $k\in\{0,1\}$ such that $k=0$ iff $\alpha$ answers $B$ and $\beta$ answers $C$:
\begin{equation}
\label{eq:nr}
   \left\{
    \begin{aligned}
     \gamma^\alpha_{j, \ell_j+1} = \gamma^\alpha_{i, \ell_i} \left(k \Theta_\kappa + (1-k) \Theta_{\kappa'} \right)+ \gamma_{j, \ell_j}^\alpha \left(k \Phi_\kappa + (1-k) \Phi_{\kappa'} \right)  \\
     \gamma_{j, \ell_j+1}^\beta = \gamma_{i, \ell_i}^\beta \left((1-k) \Theta_\kappa + k \Theta_{\kappa'} \right)+ \gamma_{j, \ell_j}^\beta \left((1-k) \Phi_\kappa + k \Phi_{\kappa'} \right) 
    \end{aligned}
        \right.
\end{equation}

\section{Propagation Patterns}
\begin{figure}[h]
\centering
\begin{subfigure}{.48\linewidth}
  \centering
  \includegraphics[width=1.\linewidth]{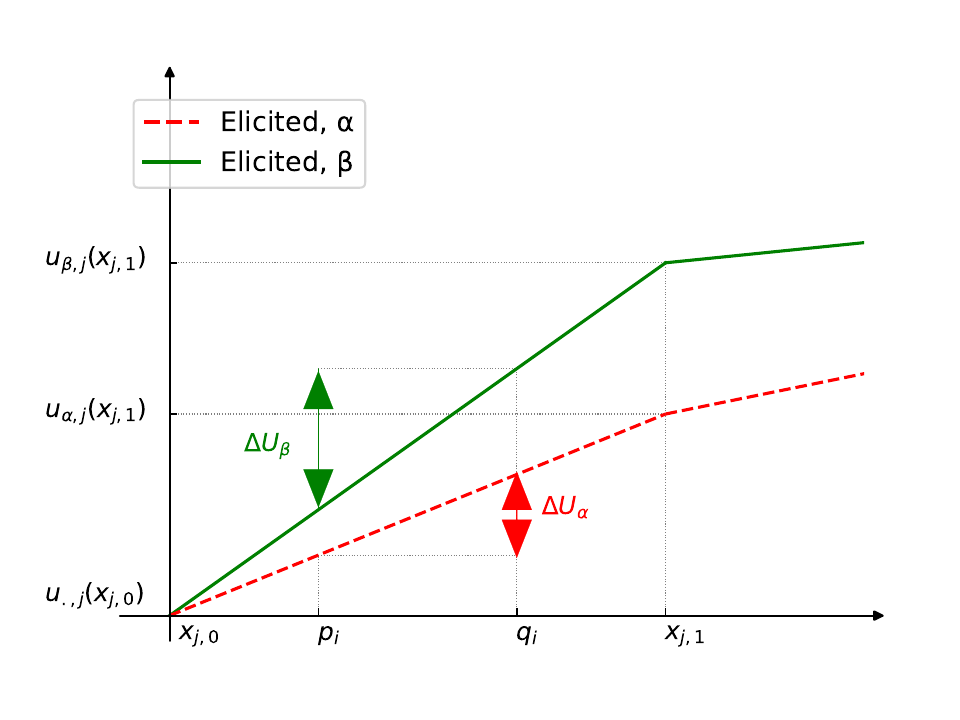}
\end{subfigure}%
\begin{subfigure}{.48\linewidth}
  \centering
  \includegraphics[width=1.\linewidth]{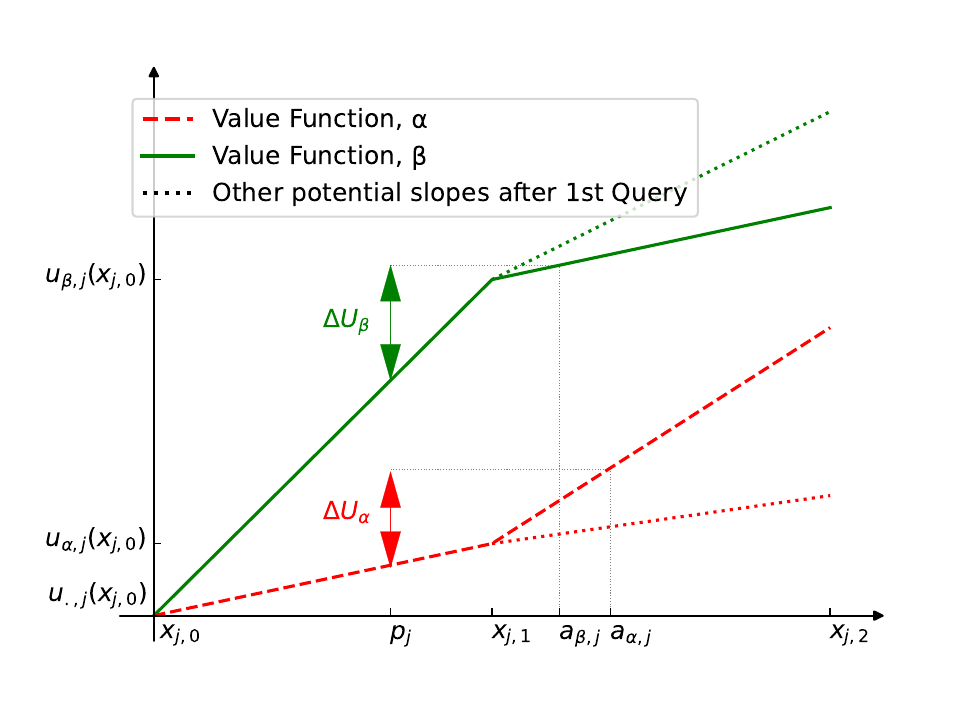}
  
\end{subfigure}
\caption{Possible value functions after Play Pattern \ref{alg:SRC}. The colored lines represent the marginal value functions of each DM.\\}
  \label{fig:pot_func}
\end{figure}

At this point, each low-level pattern introduced in Section \ref{sec:PP} yields preference information about the slopes of the indifference curves. They are exploited by introducing a binary variable $k$  representing two possible worlds per query, according to which DM provide each answer. Thus, carelessly chaining $n$ of those patterns opens $2^n$ possible worlds, as illustrated by Figure \ref{fig:pot_func}. To avoid this combinatorial explosion, we devise two higher-level sequences of moves that aim and succeed at correctly assigning the disambiguation variables $k$, effectively pruning the worlds corresponding to an incorrect assignment. These sequences differ in their requirements: in Section \ref{sub:SRSR}, we assume we know the slopes of the marginal value functions of each DM on two intervals of $\mathcal X_i$ and we obtain the slopes and their assignment to each DM on an interval of the other criterion; in Section \ref{sub:SRNR} we assume we know the slopes of the marginal value functions of each DM on an interval of $\mathcal X_i$ and an interval of $\mathcal X_j$, and we obtain the slopes and their assignment to each DM on a neighboring interval.

\subsection{Identification with two single rectangle constraints}
\label{sub:SRSR}

\begin{algorithm}
    
    \caption{Identification with two Single-Rectangle Constraints}\label{alg:2SR}
    \KwIn{A criterion $j$ and an interval labels $\ell_j \in\llbracket 1, L_j \rrbracket$; 
     a second criterion $i\ne j$, and two interval labels $\ell_i \ne \ell'_i \in\llbracket 1, L_i\rrbracket$ such that $\gamma^\alpha_{i, \ell_i}$, $\gamma^\beta_{i, \ell_i}$, $\gamma^\alpha_{i, \ell'_i}$ and $\gamma^\beta_{i, \ell'_i}$ are known.
    }
    \KwOut{the identified values of $\gamma^\alpha_{j, \ell_j}, \gamma^\beta_{j, \ell_j}$ }
    \BlankLine
    
    $A, B, C \leftarrow$ Play Pattern \ref{alg:SRC} $((i, j), (\ell_i, \ell_j))$\;\vskip.5ex
    $A', B', C' \leftarrow$ Play Pattern \ref{alg:SRC} $((i, j), (\ell'_i, \ell_j))$\;\vskip.5ex
    \vskip.5ex
    System $1a$ $\leftarrow$ Equation \ref{eq:k_t} and $A, B, C$;\vskip.5ex
    System $1b$ $\leftarrow$ Equation \ref{eq:k_t} and $A', B', C'$\;
    \vskip.5ex
    $\gamma^\alpha_{j, \ell_j}, \gamma^\beta_{j, \ell_j} \leftarrow$ Solve $(1a) = (1b)$, see Appendix \ref{app:B}\;
    \Return{$\gamma^\alpha_{j, \ell_j}, \gamma^\beta_{j, \ell_j}$}\;
\end{algorithm}

Our play patterns have been defined to be able to identify the slopes of the value functions of both DMs on a specific interval $\ell_j$ of a criterion $j$. The idea is to use two intervals $\ell_i \ne \ell'_i$ on a criterion $i$ for which the value functions are known and with $\gamma^\alpha_{ i, \ell_i}/ \gamma^\alpha_{i, \ell'_i} \ne \gamma^\beta_{i, \ell_i} / \gamma^\beta_{i, \ell'_i}$. Using Play Pattern \ref{alg:SRC}, on $\left[ x_{i, \ell_i}, x_{i, \ell_i+1}\right] \times \left[ x_{j, \ell_j}, x_{j, \ell_j+1} \right] $, we obtain a system, from Equation \ref{eq:k_t} for $\gamma_{\alpha, j, \ell_j}$ and $\gamma_{\beta, j, \ell_j}$. Doing the same for $[ x_{i, \ell'_i}, x_{i, \ell'_i+1}] \times \left[ x_{j, \ell_j}, x_{j, \ell_j} \right] $ we get a second system with different expression for these two slopes. Equating these expressions, we obtain two equations with two variables $k, k'$:
\begin{align}
\left\{
    \begin{aligned}
    \left( k \Lambda_{\kappa}  + (1-k) \Lambda_{\kappa'} \right) \gamma^\alpha_{i, \ell_i} = \left( k' \Lambda_{\kappa}'  + (1-k') \Lambda_{\kappa'}' \right) \gamma^\alpha_{i, \ell'_i} \\
    \left( (1-k) \Lambda_{\kappa}  + k \Lambda_{\kappa'} \right) \gamma^\beta_{i, \ell_i} = \left( (1-k') \Lambda_{\kappa}'  + k' \Lambda_{\kappa'}' \right) \gamma^\beta_{i, \ell_i'} \\
    \end{aligned}
\right.
    \label{eq:2SR}
\end{align}
This system is non-singular. Its resolution yields the values of variables $k$ and $k'$ and the respective slopes of the marginal value functions of each DM. Details can be found in Appendix \ref{app:B} and the described procedure is summarized in Algorithm \ref{alg:2SR}.
This result means that once one marginal value function has been identified on at least two distinct intervals, it is possible to generalize the identifiability of any interval on another criterion. 

\subsection{Identification with Single Rectangle and a Neighboring Rectangles queries}
\label{sub:SRNR}
\begin{algorithm}
    
    \caption{Identification with a Single-Rectangle and a Neighboring-Rectangles Constraints}\label{alg:SR+2R}
    \KwIn{A criterion $j$ and an interval labels $\ell_j \in\llbracket 2, L_j \rrbracket$ for which 
    $\gamma^\alpha_{j, \ell_j-1}$ and $\gamma^\beta_{j, \ell_j-1}$ are known; 
    a second criterion $i\ne j$ and an interval label $\ell_i \in\llbracket 1, L_i\rrbracket$ s.t. $\gamma^\alpha_{i, \ell_i}$ and $\gamma^\beta_{i, \ell_i}$ are known.
    }
    \KwOut{the identified values of $\gamma^\alpha_{j, \ell_j}, \gamma^\beta_{j, \ell_j}$ }
    \BlankLine
    
    $A, B, C \leftarrow$ Play Pattern \ref{alg:SRC} $((i, j), (\ell_i, \ell_j))$\;\vskip.5ex
    $A', B', C' \leftarrow$ Play Pattern \ref{alg:NRC} $((i, j), (\ell_i, \ell_j-1))$\;\vskip.5ex
    \vskip.5ex
    System $1a$ $\leftarrow$ Equation \ref{eq:k_t} and $A, B, C$;\vskip.5ex
    System $1b$ $\leftarrow$ Equation \ref{eq:nr} and $A', B', C'$\;
    \vskip.5ex
    $\gamma^\alpha_{j, \ell_j}, \gamma^\beta_{j, \ell_j} \leftarrow$ Solve $(1a) = (1b)$, see Appendix \ref{app:C}\;
    \Return{$\gamma^\alpha_{j, \ell_j}, \gamma^\beta_{j, \ell_j}$}\;
\end{algorithm}

Our second identification pattern, described in Play Pattern \ref{alg:SR+2R}, is relatively similar to the first one. The difference lies in the use of a neighboring interval of the criterion interval to be elicited instead of a second interval on the other criterion. Therefore, we will make use of one - instead of two - Single Rectangle query and one Neighboring Rectangles query. Considering an interval labeled $\ell_j-1$ on a criterion $j$ that has been elicited, we choose an additional interval $\ell_i$ on a different criterion $i$ that has also been elicited and for which 
$\gamma^\alpha_{i, \ell_i} / \gamma^\alpha_{j, \ell_j-1} \ne \gamma^\beta_{i, \ell_i} / \gamma^\beta_{j, \ell_j-1}$.
We use the Play Pattern \ref{alg:SRC}
on $\left[ x_{j, \ell_j}, x_{j, \ell_j+1}\right] \times \left[ x_{i, \ell_i}, x_{i, \ell_i+1}\right]$ to obtain the triplet $(A, B, C)$ and Play Pattern \ref{alg:NRC} on $\left[ x_{j, \ell_j-1}, x_{j, \ell_j+1}\right] \times \left[ x_{i, \ell_i}, x_{i, \ell_i+1}\right]$ to obtain the triplet $(A', B', C')$. It leads to two systems from Equations \ref{eq:k_t}, \ref{eq:nr}. These systems express different equations for $\gamma^\alpha_{j, \ell_j}$ and $\gamma^\beta_{j, \ell_j}$ and therefore can be reduced to :

\begin{align}
\left\{
    \begin{aligned}
    \left( k \Lambda_{\kappa}  + (1-k) \Lambda_{\kappa'} \right) \gamma^\alpha_{i, \ell_i} = \gamma^\alpha_{i, \ell_i} \left(k' \Theta_\kappa' + (1-k') \Theta_{\kappa'}' \right)\\+ \gamma^\alpha_{j, \ell_j-1}\left(k' \Phi_\kappa' + (1-k') \Phi_{\kappa'}' \right) \\
    \left( (1-k) \Lambda_{\kappa}  + k \Lambda_{\kappa'} \right) \gamma^\beta_{i, \ell_i} = \gamma^\beta_{i, \ell_i} \left(k' \Theta_\kappa' + (1-k') \Theta_{\kappa'}' \right)\\+ \gamma^\beta_{j, \ell_j-1}\left(k' \Phi_\kappa' + (1-k') \Phi_{\kappa'}' \right) \\
    \end{aligned}
\right.
    \label{eq:SRNR}
\end{align}
This system with two equations and two unknown variables $k, k'$ can be solved. With these values, we can finally obtain the values of $\gamma^\alpha_{j, \ell_j}$ and $\gamma^\beta_{j, \ell_j}$ using Equation \ref{eq:k_t} for example.

\section{General Elicitation Procedure}

We propose a general elicitation procedure that can be used in order to identify a set of two DMs in our described setup. The procedure can be found in the Algorithm \ref{algo:elicit} and is described in the following sections. A Python implementation and a few examples will be shared in a GitHub repository shared upon acceptance.

\RestyleAlgo{ruled}
\SetKwComment{Comment}{/* }{ */}
\SetKwBlock{initz}{\textbf{Initialize criterion $i$:}}{}
\SetKwBlock{init}{\textbf{Initialize criterion $j$: }}{}
\SetKwBlock{initi}{\textbf{Initialize criterion $j'$: }}{}
\SetKwBlock{elicit}{\textbf{Elicit Criterion $1$: }}{}
\SetKwBlock{eliciti}{\textbf{Fully Elicit Criterion $j'$: }}{}
\SetKwBlock{elicitundeux}{\textbf{Fully Elicit Criterion $i$, $j$: }}{}
\SetKwFor{For}{for}{}{}
\SetKwFor{If}{if}{}{}

\renewcommand*{\algorithmcfname}{Algorithm}

\begin{algorithm}
    \caption{General Strategy for the Elicitation}\label{alg:two}
    \initz{Set $\gamma^{\alpha}_{i, \ell_i}=\gamma^{\beta}_{i, \ell_i}=1$ for all DMs on $\left[ x_{i, \ell_i-1}, x_{i, \ell_i} \right]$} 

    \init{
    Play Pattern \ref{alg:SRC} on $\left[ x_{i, \ell_i-1}, x_{i, \ell_i} \right] \times \left[ x_{j, \ell_j}, x_{j, \ell_j+1} \right]$ \;
    Attribute slopes $\gamma^{\alpha}_{j, \ell_j}, \gamma^{\beta}_{j, \ell_j}$ on  $\left[ x_{j, \ell_j-1}, x_{j, \ell_j} \right]$ \;
  }
  \elicitundeux{
      \For{$\ell_j' \in \llbracket \ell_j, L_j \rrbracket$}{
    Play Pattern \ref{alg:SR+2R} on $\left[ x_{j, \ell_j'}, x_{j, \ell_j'+1} \right] \times \left[ x_{i, \ell_i-1}, x_{i, \ell_i} \right]$\;
    Solve Eq. \ref{eq:SRNR} and deduct slopes $\gamma^{\alpha}_{j, \ell_j'}, \gamma^{\beta}_{j, \ell_j'}$\;
    }
      \For{$\ell_i' \in \llbracket \ell_i, L_i \rrbracket$}{
    Play Pattern \ref{alg:SR+2R} on $\left[ x_{i, \ell_i'}, x_{i, \ell_i'+1} \right] \times \left[ x_{j, \ell_j-1}, x_{j, \ell_j} \right]$\;
    Solve Eq. \ref{eq:SRNR} and deduct slopes $\gamma^{\alpha}_{i, \ell_i'}, \gamma^{\beta}_{i, \ell_i'}$\;
    }
  }
\For{criterion $j' \in \mathcal{N}, j \ne i, j$}{
    \initi{
    Play Pattern \ref{alg:2SR} on $\left[ x_{j', 0}, x_{j', 1} \right] \times \left[ x_{i, \ell_i}, x_{i, \ell_i+1} \right]$ \;
    Solve Eq. \ref{eq:2SR} and deduct slopes $\gamma^\alpha_{j', 1}, \gamma^\beta_{j', 1}$ \;
    }
      \eliciti{
    \For{$\ell_{j'} \in \llbracket 1, L_{j'}\rrbracket$}{
    Play Pattern \ref{alg:SR+2R} on $\left[ x_{j', \ell_{j'}}, x_{j', \ell_j'+1} \right] \times \left[ x_{i, \ell_i}, x_{i, \ell_i+1} \right]$\;
    Solve Eq. \ref{eq:SRNR} and deduct slopes $\gamma^\alpha_{j', \ell_j'}, \gamma^\beta_{j', \ell_j'}$\;
    }
  }
}
\label{algo:elicit}

\end{algorithm}

\subsection{Initialization}

We begin by choosing two criteria $i\ne j$ and a rectangle $\mathcal R_{\ell_i,\ell_j}\subset\mathcal X_i \times \mathcal X_j$ and following Play Pattern 1 so as to obtain Single Rectangle preference information describing two \emph{intersecting} indifference curves (i.e. distinct points $B$ and $C$). If this is not possible, the DMs have the exact same preferences and the identification task is over. Otherwise, we set:
$$ \gamma^\alpha_{i,\ell_i}=1 ;\quad \gamma^\alpha_{j, \ell_j} = \frac{B_i - A_i}{A_j - B_j};\quad \gamma^\beta_{i,\ell_i}=1 ; \quad \gamma^\beta_{j, \ell_j} = \frac{C_i - A_i}{A_j - C_j}$$
Indeed, without loss of generality, we can arbitrarily assign the answer $B$ to the DM $\alpha$ (resp. $C$ to $\beta$) and normalize the UTA model of $\alpha$ (resp. $\beta$) so that it has unit slope on $[x_{i,\ell_i-1},x_{i,\ell_i}]$ \footnote{This normalization is licit but not standard in MAVT. The usual convention of having values ranging from 0 to 1 can be easily restored after the complete identification of the two AVF.}.

In order to find a suitable rectangle, we traverse the intervals in $\mathcal X_i$ and $\mathcal X_j$ in increasing order. Thus, when one is found, all rectangles with lower indices have unanimous DMs and can be identified at this point.

\subsection{Full elicitation of criteria $i$ and $j$}
\label{sec:full_elicit}
It is now possible to use the Play Pattern \ref{alg:SR+2R} to elicit interval by interval the criteria $i$ and $j$, in increasing order.
Let us consider that the criterion $i$ has been elicited up to the $\ell$-th interval, we formulate: 
\begin{itemize}
    \item a Single Rectangle Query on $\left[ x_{j, \ell_j-1}, x_{j, \ell_j} \right] \times \left[ x_{i, \ell+1}, x_{i, \ell+2} \right]$
    \item a Neighboring Rectangles Query on $
    \left[ x_{j, \ell_j-1}, x_{j, \ell_j} \right]
    \times \left[ x_{i, \ell}, x_{i, \ell+2} \right]$
\end{itemize}
Following Equation \ref{eq:SRNR}, we are able to obtain the slopes values $\gamma^\alpha_{i, \ell+1}$ and $\gamma^\beta_{i, \ell+1}$. Using this procedure iteratively for $2 \leq \ell \leq L_i$, it is possible to fully elicit the criterion $i$.

\begin{figure}[h]
    \centering
 \includegraphics[width=0.4\textwidth]{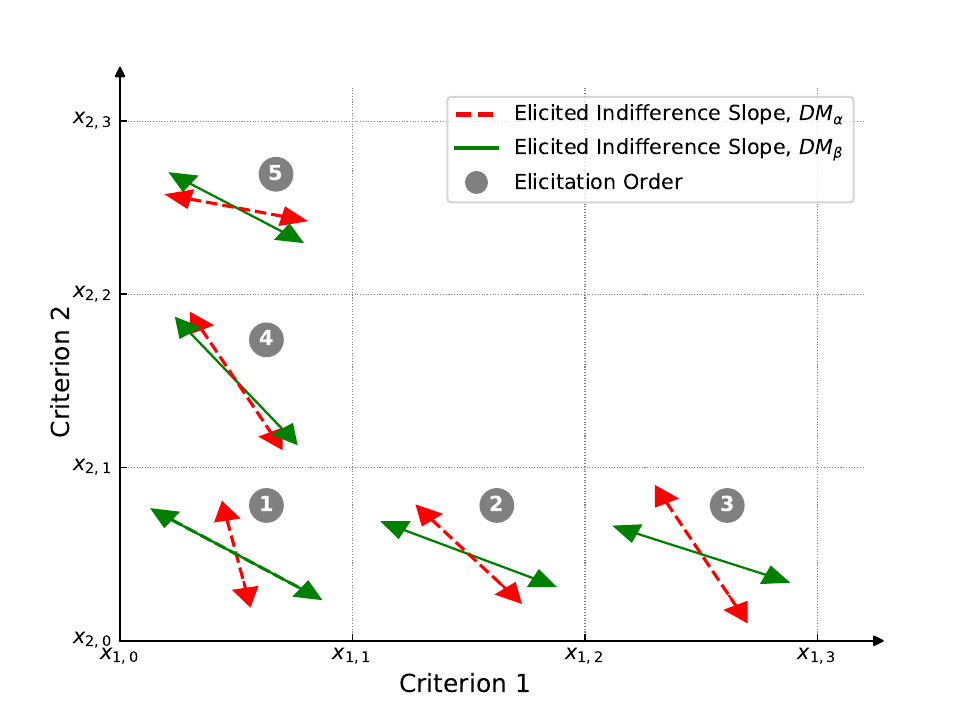}
\caption{General Procedure sequencing for the elicitation of two different DMs on two criteria.}
\label{fig:general}
\vskip3ex
\end{figure}

The same procedure can be applied for criterion $j$. The only change needed is the use of the interval $\left[ x_{j, \ell_j-1}, x_{j, \ell_j} \right]$ 
instead of $\left[ x_{i, \ell_i-1}, x_{i,\ell_i} \right]$ 
for the different queries. The iterative steps for two criteria are illustrated in Figure \ref{fig:general}, assuming initialization occurs in the rectangle $[x_{i,0},x_{i,1}]\times[x_{j,0},x_{j,1}]$. Observe it is necessary and sufficient to  exploit preference information concerning exactly one rectangle per column and per row (i.e. $L_i+L_j-1$ rectangles) to fully identify $u^\alpha_i, u^\alpha_j, u^\beta_i, u^\beta_j$. 

\subsection{Elicitation of the remaining criteria}
The remaining criteria can finally be elicited independently. For a criterion $j' \notin\{i,j\}$, an initialization is done in order to get the slopes on the first interval, following the Play Pattern \ref{alg:2SR}:
\begin{itemize}
    \item a Single Rectangle Query on $\left[ x_{i, \ell_i-1}, x_{i, \ell_i} \right] \times \left[ x_{j', 0}, x_{j', 1} \right]$
    \item a Single Rectangle Query on $\left[ x_{i, \ell_i-1}, x_{i, \ell_i} \right] \times \left[ x_{j', 0}, x_{j', 1} \right]$
\end{itemize}

Following Equation \ref{eq:2SR}, we obtain the slopes values $\gamma^\alpha_{i, 1}$ and $\gamma^\beta_{i, 1}$.

At this point, it is possible to use the procedure for criterion $i$ or $j$ described in Section \ref{sec:full_elicit} to fully elicit the criterion $j'$.

\section*{Conclusion}
In this paper, we study the identifiability of two additive piecewise linear value functions using matching queries. Such queries provide two answers (those of the two additive models), but without specifying to which model each answer corresponds. We provide a systematic elicitation procedure that makes it possible to obtain the two additive piecewise linear models from a finite sequence of matching queries, hence proving identifiability. 

This work leaves open questions that should be studied in further research. First, our work is limited to additive value functions for which the marginals are piecewise linear; it would be interesting to study this identifiability problem when marginals can be any monotonically increasing function. A second extension relates to the fact that our work accounts for two DMs only; extending Single- and Neighboring-Rectangle  queries to more than two DMs seems straightforward, but defining an elicitation procedure to identify $n$ additive models ($n>2$) would be of great interest. Finally, the positive results concerning the identifiability of such model could help to analyze the behavior of statistical learning algorithms.






\bibliography{mybibfile}

\newpage
\clearpage

\appendix

\begin{figure*}
\section{Details for Equation \ref{eq:k_t}}
\begin{align*}
    \left\{
\begin{aligned}
    (A_i, A_j)\sim_{\kappa} (B_i, B_j)\\
    (A_i, A_j)\sim_{\kappa'} (C_i, C_j)\\
\end{aligned}
    \right.
\Leftrightarrow &
    \left\{
\begin{aligned}
    u_{\kappa}((A_i, A_j)) = u_{\kappa}((B_i, B_j))\\
    u_{\kappa'}((A_i, A_j)) = u_{\kappa'}((C_i, C_j))
\end{aligned}
    \right.
\end{align*}
Since $A_i, B_i, C_i \in \left[x_{i, \ell_i}, x_{i, \ell_i+1} \right]$ and $A_j, B_j, C_j \in \left[x_{i, \ell_j}, x_{j, \ell_j+1} \right]$:
\begin{align*}
    & 
    \left\{
    \begin{aligned}
        &u_{i}^\kappa(x_{i, \ell_i}) + \gamma^\kappa_{i, \ell_i+1} \cdot (A_i - x_{i, \ell_i}) +  u^\kappa_{j}(x_{j, \ell_j}) + \gamma^\kappa_{j, \ell_j+1} \cdot (A_j - x_{j, \ell_j}) = u_{i}^\kappa(x_{i, \ell_i}) + \gamma^\kappa_{i, \ell_i+1} \cdot (B_i - x_{i, \ell_i}) +  u^\kappa_{j}(x_{j, \ell_j}) + \gamma^\kappa_{j, \ell_j+1} \cdot (B_j - x_{j, \ell_j})\\
        &u_{i}^{\kappa'}(x_{i, \ell_i}) + \gamma^{\kappa'}_{i, \ell_i+1} \cdot (A_i - x_{i, \ell_i}) +  u^{\kappa'}{j}(x_{j, \ell_j}) + \gamma^{\kappa'}_{j, \ell_j+1} \cdot (A_j - x_{j, \ell_j}) = u_{i}^{\kappa'}(x_{i, \ell_i}) + \gamma^{\kappa'}_{i, \ell_i+1} \cdot (C_i - x_{i, \ell_i}) +  u^{\kappa'}{j}(x_{j, \ell_j}) + \gamma^{\kappa'}_{j, \ell_j+1} \cdot (C_j - x_{j, \ell_j})\\
    \end{aligned}
    \right.\\
    \Leftrightarrow & \left\{
    \begin{aligned}
        &\gamma^\kappa_{i, \ell+1} \cdot (A_i - B_i) + \gamma^\kappa_{j, \ell+1} \cdot (A_j - B_j) = 0\\
        &\gamma^{\kappa'}_{i, \ell_i+1} \cdot (A_i - C_i) + \gamma^{\kappa'}_{j, \ell_j+1} \cdot (A_j - C_j) = 0\\
    \end{aligned}
    \right.\\
    \Leftrightarrow & \text{Equation } \ref{eq:k_t}
\end{align*}
\label{app:eq_2}
\end{figure*}

\begin{figure*}
\section{Condition of unicity of the system resulting from Play Pattern \ref{alg:2SR}}
\label{app:B}
Using Play Pattern \ref{alg:SRC}, on $\left[ x_{i, \ell_i}, x_{i, \ell_i+1}\right] \times \left[ x_{j, \ell_j}, x_{j, \ell_j+1} \right] $, we obtain a system, from Equation \ref{eq:k_t}:
\begin{equation}
        \left\{
    \begin{aligned}
    &\gamma^\alpha_{j, \ell_j} = \left( k \cdot \Lambda_{\kappa} + (1-k) \cdot \Lambda_{\kappa'} \right) \cdot \gamma^\alpha_{i, \ell_i}\\
    &\gamma^\beta_{j, \ell_j} = \left( (1-k) \cdot \Lambda_{\kappa}  + k \cdot \Lambda_{\kappa'} \right) \cdot \gamma_{i, \ell_i}^\beta\\
    \end{aligned}
        \right.
\end{equation}

Following the same procedure with $\left[ x_{i, \ell'_i}, x_{i, \ell'_i+1}\right] \times \left[ x_{j, \ell_j}, x_{j, \ell_j} \right] $ we get a second system with different expression for our two slopes:
\begin{equation}
        \left\{
    \begin{aligned}
    &\gamma^\alpha_{i, \ell_i} = \left( k' \cdot \Lambda_{\kappa'}'  + (1-k') \cdot \Lambda_{\kappa'}' \right) \cdot \gamma^\alpha_{ j, \ell_j}\\
    &\gamma^\beta_{i, \ell_i} = \left( (1-k') \cdot \Lambda_{\kappa}'  + k' \cdot \Lambda_{\kappa'}' \right) \cdot \gamma_{j, \ell_j}^\beta\\
    \end{aligned}
        \right.
\end{equation}

Equating the two systems, we obtain two equations with two variables $k, k'$, which we can solve:
\begin{align}
    & \left\{
    \begin{aligned}
    \left( k \Lambda_{\kappa}  + (1-k) \Lambda_{\kappa'} \right) \gamma^\alpha_{i, \ell_i} = \left( k' \Lambda_{\kappa}'  + (1-k') \Lambda_{\kappa'}' \right) \gamma^\alpha_{i, \ell'_i} \\
    \left( (1-k) \Lambda_{\kappa}  + k \Lambda_{\kappa'} \right) \gamma^\beta_{i, \ell_i} = \left( (1-k') \Lambda_{\kappa}'  + k' \Lambda_{\kappa'}' \right) \gamma^\beta_{i, \ell_i'} \\
    \end{aligned}
    \right.\\
    \Leftrightarrow & \left\{
    \begin{aligned}
        \left[ k \left( \Lambda_\kappa - \Lambda_{\kappa'} \right) + \Lambda_{\kappa'} \right] \gamma^\alpha_{i, \ell_i} = \left[ k \left( \Lambda_\kappa' - \Lambda_{\kappa'}' \right) + \Lambda_{\kappa'}' \right] \gamma^\alpha_{i, \ell'_i}\\
        \left[ k \left( \Lambda_{\kappa'} - \Lambda_{\kappa} \right) + \Lambda_{\kappa} \right] \gamma^\beta{i, \ell_j} = \left[ k \left( \Lambda_{\kappa'}' - \Lambda_{\kappa}' \right) + \Lambda_{\kappa}' \right] \gamma^\beta\\
    \end{aligned}
    \right.
\end{align}

If the equations are linearly correlated, it means that the coefficients in $k, k'$ are correlated between the equations. Differently said, it would mean that $\exists z \in \mathbb{R}$ such that:

\begin{align*}
&        \left\{
    \begin{aligned}
    &(\Lambda_{\kappa}  - \Lambda_{\kappa'}) \gamma^\alpha_{i, \ell_i} = z \cdot (\Lambda_{\kappa'}  - \Lambda_{\kappa}) \gamma^\beta_{i, \ell_i}\\
    &(\Lambda'_{\kappa'}  - \Lambda'_{\kappa}) \gamma^\beta{i, \ell_i'} = z \cdot (\Lambda'_{\kappa}  - \Lambda'_{\kappa'}) \gamma^\alpha_{i, \ell_i'}\\
    \end{aligned}
        \right.\\
\Leftrightarrow &        \left\{
    \begin{aligned}
    &\gamma^\alpha_{i, \ell_i} = - z \cdot  \gamma^\beta_{i, \ell_i}\\
    &\gamma^\beta_{i, \ell_i'} = - z \cdot \gamma^\alpha_{i, \ell_i'}\\
    \end{aligned}
        \right.\\
\end{align*}

The second system holds if the $\Lambda$ values are different at $t$ and $t+1$. Which is the case if the answers of the two DMs are different.
It results that the solution of the system can be found and is unique unless:
\begin{itemize}
    \item the answers from the query $t$ or $t'$ are the same for both DMs
    \item the slopes ratios of each DM are the same on the two intervals on criterion $i$, $\frac{\gamma^\alpha_{i, \ell_i}}{\gamma^\alpha_{i, \ell_i'}} = \frac{\gamma^\beta_{i, \ell_i}}{\gamma^\beta_{i, \ell_i'}}$
\end{itemize}
\end{figure*}

\begin{figure*}
\section{Condition of unicity of the system resulting from Play Pattern \ref{alg:SR+2R}}
Using Play Pattern \ref{alg:SRC}, on $\left[ x_{i, \ell_i}, x_{i, \ell_i+1}\right] \times \left[ x_{j, \ell_j}, x_{j, \ell_j+1} \right] $, we obtain a system, from Equation \ref{eq:k_t}:
\begin{equation}
        \left\{
    \begin{aligned}
    &\gamma^\alpha_{j, \ell_j+1} = \left( k \cdot \Lambda_{\kappa}  + (1-k) \cdot \Lambda_{\kappa'} \right) \cdot \gamma^\alpha_{i, \ell_i}\\
    &\gamma^\beta_{j, \ell_j+1} = \left( (1-k) \cdot \Lambda_{\kappa}  + k \cdot \Lambda_{\kappa'} \right) \cdot \gamma_{i, \ell_i}^\beta\\
    \end{aligned}
        \right.
\end{equation}

Following Play Pattern \ref{alg:NRC} with $\left[ x_{i, \ell_i}, x_{i, \ell_i+1}\right] \times \left[ x_{j, \ell_j-1}, x_{j, \ell_j+1} \right] $ we get a second system from Equation \ref{eq:nr}
\begin{equation}
       \left\{
    \begin{aligned}
     \gamma^\alpha_{j, \ell_j+1} = \gamma^\alpha_{i, \ell_i} \left(k' \Theta_\kappa + (1-k') \Theta_{\kappa'} \right)+ \gamma_{j, \ell_j}^\alpha \left(k' \Phi_\kappa + (1-k') \Phi_{\kappa'} \right)  \\
     \gamma_{j, \ell_j+1}^\beta = \gamma_{i, \ell_i}^\beta \left((1-k') \Theta_\kappa + k' \Theta_{\kappa'} \right)+ \gamma_{j, \ell_j}^\beta \left((1-k') \Phi_\kappa + k' \Phi_{\kappa'} \right)  \\
    \end{aligned}
        \right.\\
\end{equation}

Equating the two systems, and following Appendix \ref{app:B}, we find that if the equations are linearly correlated, it means that the coefficients in $k, k'$ are correlated between the equations. Differently said, it would mean that $\exists z \in \mathbb{R}$ such that:

\begin{align*}
&        \left\{
    \begin{aligned}
    &(\Lambda_{\kappa}  - \Lambda_{\kappa'}) \gamma^\alpha_{i, \ell_i} = z \cdot (\Lambda_{\kappa'}  - \Lambda_{\kappa}) \gamma^\beta_{i, \ell_i}\\
    &(\Theta'_\kappa -\Theta'_{\kappa'} )\gamma^\alpha_{i, \ell_i} + (\Phi'_{\kappa}  - \Phi'_{\kappa'}) \gamma^\alpha_{j, \ell_j} = z \cdot \left[(\Theta'_{\kappa'} -\Theta'_{\kappa} )\gamma^\beta_{i, \ell_i} + (\Phi'_{\kappa'}  - \Phi'_{\kappa}) \gamma^\beta_{j, \ell_j} \right]\\
    \end{aligned}
        \right.\\
\Leftrightarrow &        \left\{
    \begin{aligned}
    &\gamma^\alpha_{i, \ell_i} = - z \cdot  \gamma^\beta_{i, \ell_i}\\
     &\gamma^\alpha_{j, \ell_j} = z \cdot \gamma^\beta_{j, \ell_j}\\
    \end{aligned}
        \right.\\
\end{align*}

Similarly to \ref{app:B}, we find that the solution of the system is unique unless:
\begin{itemize}
    \item the answers from the query $t$ or $t'$ are the same for both DMs
    \item the slopes ratios of each DM are the same on the two intervals on criterion $j$, $\frac{\gamma^\alpha_{i, \ell_i}}{\gamma^\beta_{i, \ell_i}} = \frac{\gamma^\alpha_{j, \ell_j}}{\gamma^\beta_{j, \ell_j}}$
\end{itemize}

\label{app:C}
\end{figure*}
\end{document}